\documentclass{article} 
\usepackage{nips14submit_e,times}
\usepackage{url}
\usepackage{graphicx}
\usepackage{subfig}
\usepackage{amsfonts}
\usepackage{multirow}

\usepackage{algorithmic}
\usepackage{algorithm}

\title{Non-Convex Boosting Overcomes Random Label Noise}

\author{
Sunsern Cheamanunkul\\
Department of Computer Science and Engineering\\
University of California, San Diego\\
La Jolla, CA 92093 \\
\texttt{scheaman@eng.ucsd.edu} \\
\And
Evan Ettinger \\
Department of Computer Science and Engineering\\
University of California, San Diego\\
La Jolla, CA 92093 \\
\texttt{evanettinger@gmail.com}\\
\And
Yoav Freund \\
Department of Computer Science and Engineering\\
University of California, San Diego\\
La Jolla, CA 92093 \\
\texttt{yfreund@eng.ucsd.edu}
}

\nipsfinalcopy 

\newcommand{\newmcommand}[2]{\newcommand{#1}{{\ifmmode {#2}\else\mbox{${#2}$}\fi}}}
\newcommand{\renewmcommand}[2]{\renewcommand{#1}{{\ifmmode {#2}\else\mbox{${#2}$}\fi}}}
\newcommand{\newmcommandi}[2]{\newcommand{#1}[1]{{\ifmmode {#2}\else\mbox{${#2}$}\fi}}}
\newcommand{\newmcommandii}[2]{\newcommand{#1}[2]{{\ifmmode {#2}\else\mbox{${#2}$}\fi}}}
\newcommand{\newmcommandiii}[2]{\newcommand{#1}[3]{{\ifmmode {#2}\else\mbox{${#2}$}\fi}}}

\newcommand{\eqref}[1]{Eq.~(\ref{#1})}

\newfont{\msym}{msbm10}

\newcommand{\paren}[1]{\left({#1}\right)}
\newcommand{\brackets}[1]{\left[{#1}\right]}

\newcommand{\ceil}[1]{\left\lceil{#1}\right\rceil}

\newcommand{\sfrac}[2]{\mbox{$\frac{#1}{#2}$}}
\newcommand{\half}{\sfrac{1}{2}}
\newcommand{\sign}{{\rm sign}}

\newcommand{\floor}[1]{\left\lfloor{#1}\right\rfloor}

\newcommand{\hf}{H}  

\newcommand{\hyp}{{\cal H}} 

\newmcommand{\M}{\bf M}
\newmcommand{\dM}{\M'}
\newmcommand{\D}{{\cal D}}
\renewmcommand{\P}{P}
\newmcommand{\Q}{Q}
\newmcommand{\Pt}{\P_t}
\newmcommand{\Qt}{\Q_t}
\newmcommand{\Pstar}{\P^*}
\newmcommand{\Pref}{\tilde{\P}}	
\newmcommand{\Qstar}{\Q^*}
\newmcommand{\Pa}{\overline{\P}}
\newmcommand{\Qa}{\overline{\Q}}
\newmcommandi{\trans}{{#1}^{\top}}
\newmcommand{\mhx}{\M(h,x)}
\newmcommand{\mxh}{\dM(x,h)}
\newmcommand{\mpq}{\M(\P,\Q)}
\newmcommand{\mpsq}{\M(\Pstar,\Q)}
\newmcommand{\mpsqt}{\M(\Pstar,\Qt)}
\newmcommand{\mptqt}{\M(\Pt,\Qt)}
\newmcommand{\mptq}{\M(\Pt,\Q)}
\newmcommand{\mpqt}{\M(\P,\Qt)}

\newmcommand{\sumt}{\sum_{t=1}^T}
\newmcommand{\prodt}{\prod_{t=1}^T}
\newmcommand{\sumim}{\sum_{i=1}^m}
\newmcommand{\delt}{\Delta_T}

\newmcommand{\hyps}{\hyp}
\newmcommand{\predt}{\hat{y}_t}
\newmcommandii{\prob}{\Pr_{#1}\brackets{{#2}}}

\newmcommand{\hfin}{\hf}

\newcommand{\rpot}[2]{\Phi(#1,#2)}

\newcommand{\s}{s}

\newcommand{\erf}{\mbox{erf}}

\newcommand{\bc}{\beta} 

\newlength{\axiswd}  
\newlength{\axisht}  
\newlength{\boxwd}   
\newlength{\boxht}   
\newlength{\terminwidth}  

\newcommand{\termout}[1]{ }

\begin{document}

\maketitle

\begin{abstract}
  The sensitivity of Adaboost to random label noise is a well-studied
  problem. LogitBoost, BrownBoost and RobustBoost are boosting
  algorithms claimed to be less sensitive to noise than AdaBoost.
  We present the results of experiments evaluating these algorithms on
  both synthetic and real datasets. We compare the performance on each
  of datasets when the labels are corrupted by different levels of
  independent label noise. In presence of random label noise, we
  found that BrownBoost and RobustBoost perform significantly better
  than AdaBoost and LogitBoost, while the difference between each pair
  of algorithms is insignificant.  We provide an explanation for the
  difference based on the margin distributions of the algorithms.
\end{abstract}

\section{Introduction} \label{sec:intro}
Adaboost~\cite{Schapire2012} is a very popular classification learning
algorithm. It is a simple and effective algorithm. While generally
successful, the sensitivity of Adaboost to random label noise is well
documented~\cite{Freund1996, Maclin1997a, Dietterich2000}.  The random
label noise setup is one where we take a dataset for which our
learning algorithm generates an accurate classifier and we flip each
label in the training set with some small fixed probability. Note that
the classifier that was a good classifier in the noiseless setup is
still a good classifier. The problem is that in the noisy setup the
noisy examples mislead the learning algorithm and cause it to diverge
significantly from the good classifier.

LogitBoost~\cite{Friedman1998} is believed to be less sensitive to
random noise than Adaboost, but it still falls pray to high levels of
random labels noise.

In fact, Servedio and Long~\cite{Long2008} proved that, in general,
any boosting algorithm that uses a convex potential function can be
misled by random label noise. Freund~\cite{Freund2001} suggested a
boosting algorithm, called Brownboost, that uses a {\em non-convex}
potential function and claims to overcome random label noise. The main
contribution of this paper is experimental evidence that support this
claim. The other contribution is a heuristic for automatically tuning
the parameters that Brownboost needs as input.

\section{Boosting, margins and convexity}
All non-recursive boosting algorithms generate a classification rule
which is a thresholded linear combination of so-called ``base''
classification rules. More precisely, let $(x,y)$, with $y
\in\{-1,+1\}$ denote a labeled example. Let $h_i
:X \to \{-1,+1\}$ denote
the base rules. then the output of the
boosting algorithm is a rule of the form
\[
F(x) = \sign\left( \sum_i \alpha_i h_i(x) \right)
\]
As it turns out, the sum which is the operand of the $\sign$ function
is important for understanding the operation of boosting algorithms as
well as the generalization error of the generated classifier. It is
convenient to replace the sum with a dot product: 
$$ \sum_i \alpha_i h_i(x) = \vec{\alpha}\cdot\vec{h}(x)$$
with $\vec{\alpha}$ and $\vec{h}$ defined in the natural way.

To characterize the relationship of the value of the sum and the label
$y$, Schapire et. al.~\cite{Schapire1998} defines the ``margin'' of an
example as: $$m(x,y) = y \vec{\alpha}\cdot\vec{h}(x)$$
Thus $m(x,y)>0$ if and only if the classification rule is correct on
the example $(x,y)$. The natural goal is therefore to find base rules
$\{h_i\}$ and weights $\{\alpha_i\}$ such that the number of training
examples with negative margin i.e. the number of misclassified
examples is minimized.

{\bf From a computational point of view,} the easy case occurs when the
training data is {\em separable}. In other words, when when there
exists a setting of $\alpha$ such that $m(x,y)>0$ for all of the
training examples. In that case {\em finding} an appropriate setting
for $\alpha$ is easy and can be done using the perceptron algorithm.

On the other hand, when the training set is not linearly separable,
the problem of finding the error minimizing plane is NP-hard. We
therefore have to resort to approximations. The approximation used in
Adaboost and Logitboost is to use a convex function that upper bounds
the step function that corresponds to the number of
misclassifications. Specifically, Adaboost corresponds to minimizing
the potential function:
\[
\phi(x,y)=e^{-m(x,y)}
\]
and Logitboost corresponds to minimizing the potential function
\[
\phi(x,y)=\ln \left(1+e^{-m(x,y)}\right)
\]
As both of these potential functions are convex, minimizing them can
be done efficiently. See figure~1 for a depiction of the 0/1 error
function and of the potential functions corresponding to Adaboost and
LogitBoost. Using a convex upper bound makes the problem tractable,
but obviously there can be a significant gap between the bound and the
step function which can lead us to a sub-optimal solution.

Moreover, as was shown in~\cite{Long2008}, algorithms that minimize
convex potential functions can {\em always} be fooled by the addition
of random label noise. This naturally leads us to considering non-convex
potential functions that upper bound the 0/1 error function. However,
before we get to that. We devote a section to the question: ``is
minimizing the training error the right goal for a learning algorithm?''

\subsection{Margins and generalization}
Our ultimate goal when learning classifiers to reduce the test error -
the number of mistakes the classifier makes on the test set. As we
only have access to the training data we cannot minimize the
generalization error directly. The natural goal of the algorithm is to
minimize the {\em training} error.  However, Schapire
et. al.~\cite{Schapire1998}, showed that there is a better performance
measure the performance of the
boosted classifier on the training set. That is to maximize the number
of training examples whose normalized margin is larger than some
$\theta>0$. Where the positive margin of the example $(x,y)$ is
defined to be \newcommand{\nm}{\hat{m}}
$$\nm(x,y) = y \frac{\vec{\alpha}\cdot\vec{h}(x)}{\|\alpha\|_1}$$
The intuition, presented and justified in~\cite{Schapire1998}, is that
large positive margins correspond to confident
predictions. Specifically, Theorem~2 in~\cite{Schapire1998} states
that, with probability $1-\delta$ over the random choice of the
training set, the following inequality holds for all $\theta>0$
\begin{equation} \label{eqn:MarginsTheorem}
P_D\left(\nm(x,y) \le 0\right)
\leq
P_S\left(\nm(x,y) \le \theta\right)
+ O\left(\frac{1}{\sqrt{n}}
\left(\frac{d\log^2(n/d)}{\theta^2} + \log(1/\delta)\right)^{1/2}
\right).
\end{equation}
where $P_D$ is the probability with respect to the true
distribution, $P_S$ is the probability with respect to the training
set $S$ whose size is $n$ and $d$ is the VC dimension of the base
classifiers.

Note that the bound consists of two terms, the first corresponds to the
fraction of the training set whose margin is at most $\theta$ and the
second which is $O(1/\theta)$. The first term increases with $\theta$
while the second term decreases with $\theta$. As the bound holds
uniformly for all values of $\theta$ we are free to choose the values
of $\theta$ that would minimize the bound. Intuitively, the first term
corresponds to the examples on which we ``give up''. Note that giving
up on an example increases the RHS of
Equation~\ref{eqn:MarginsTheorem} by $1/m$ regardless of amount by
which the margin of the example is smaller than $\theta$.

\begin{figure}[t]
  \centering
  \includegraphics[width=0.45\textwidth]{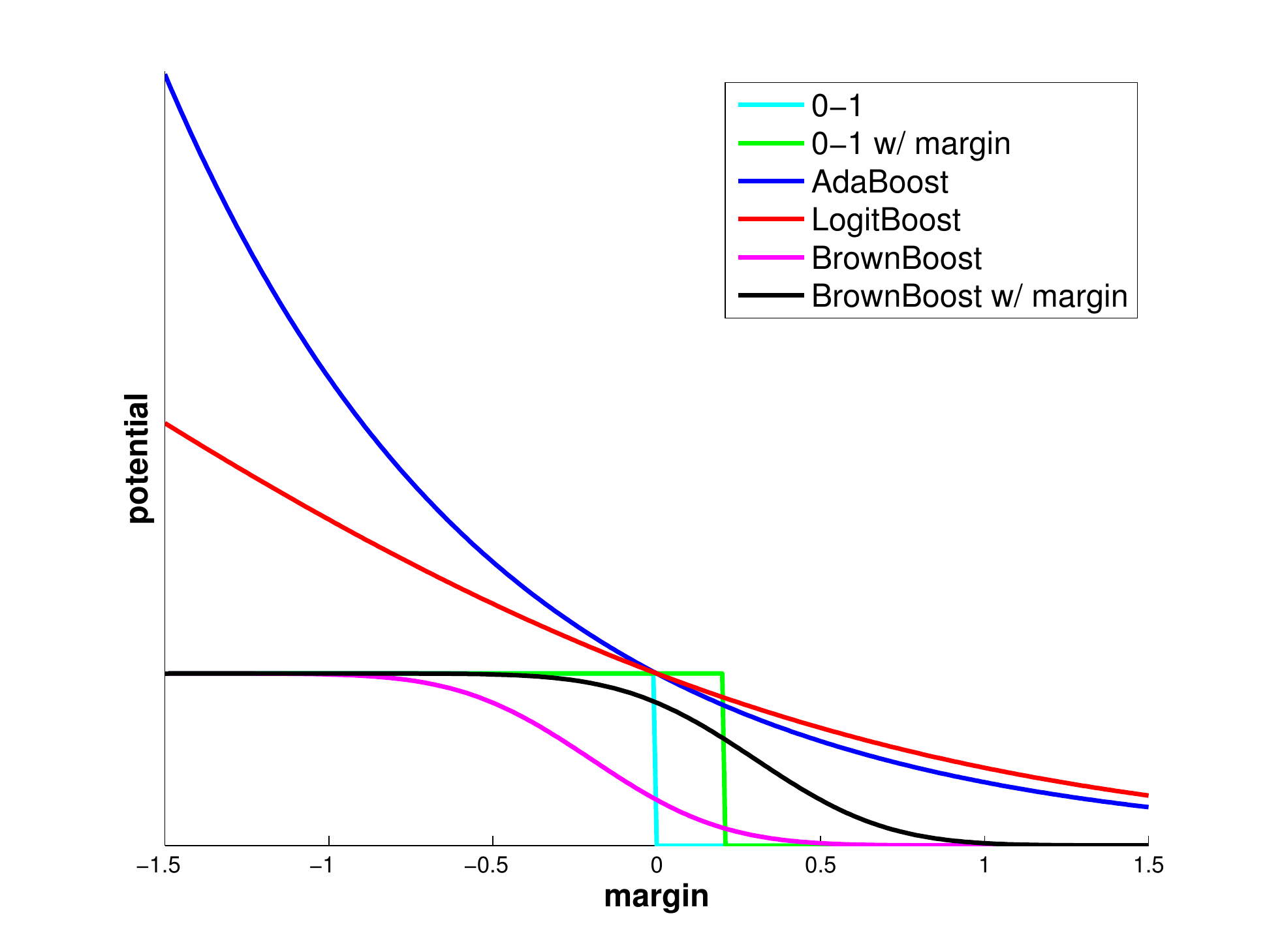}
  \includegraphics[width=0.45\textwidth]{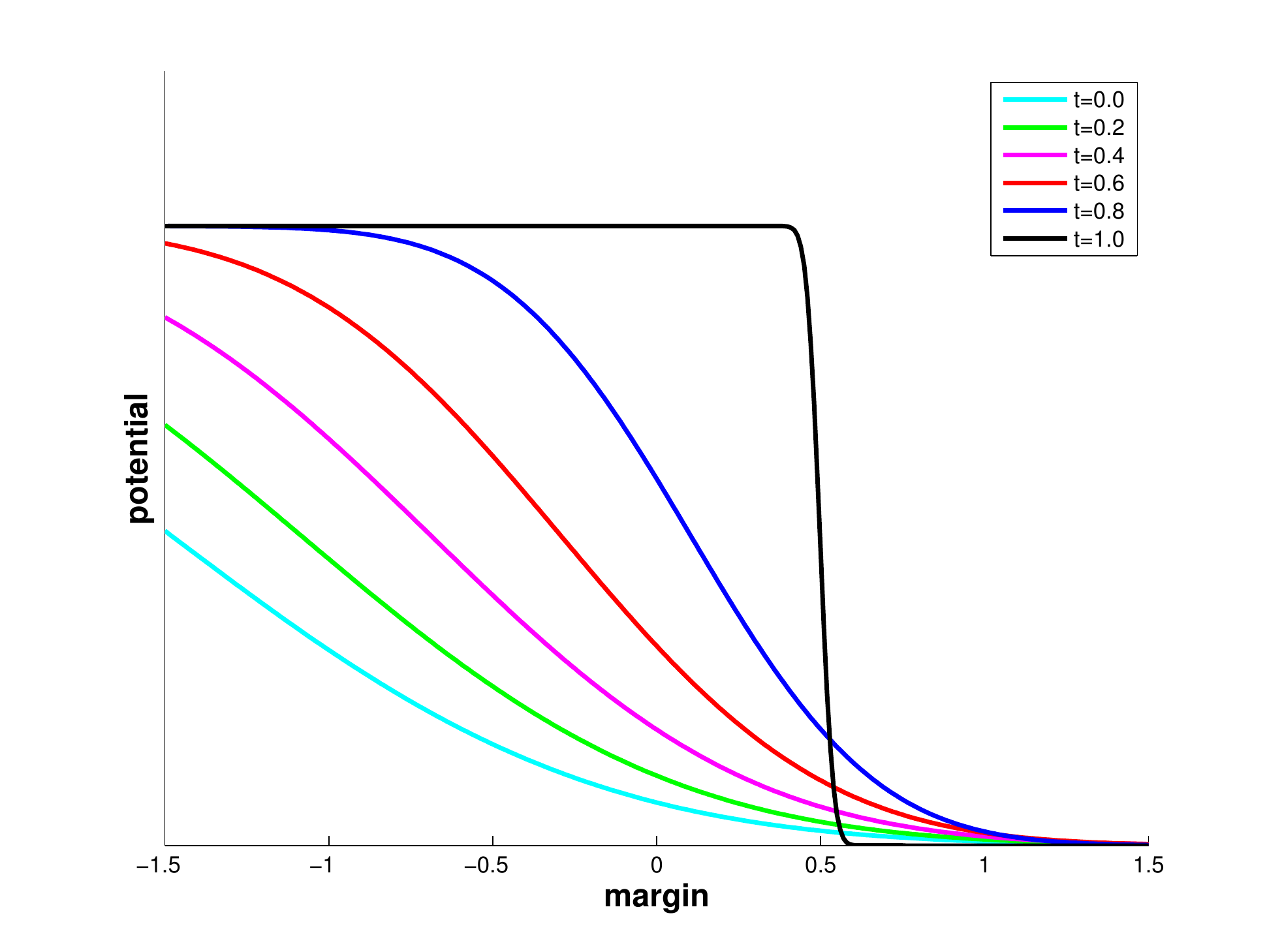}
  \caption{Left: Zero/one loss function with and without margin
    plotted with potentials for AdaBoost, LogitBoost, confidence rated
    BrownBoost and confidence rated RobustBoost. Right: How the
    potentials for BrownBoost vary with the time
    parameter $t$.~\label{fig:potentials}}
\end{figure}

The goal of learning now becomes to minimize a step function that is
thresholded at $\theta$ (see Figure~\ref{fig:potentials}). This does not make the problem
any easier than minimizing the training error. The suggestion is,
however, that minimizing this potential function will yield
classifiers with smaller test error.

\subsection{Random label noise}
Consider the effect of random label noise on boosting.  The weight
assigned to example $(x,y)$ by Adaboost is exponential in the margin
$w(x,y) = e^{-m(x,y)}$. Suppose $c(x)$ is the best rule for the
noise-free data. Suppose we now add independent label noise to the
dataset. The margin of $c(x)$ on examples whose label has been flipped
will be negative, resulting in a large weight being assigned to the
noisy examples, resulting in base classifiers that fit the noisy
examples.

\section{Three non-convex boosting algorithms}
Freund~\cite{Freund1995,Freund2001,Freund2009} suggested several boosting
algorithms that use non-convex potential functions. We briefly review
three of these algorithms: Boost-by-majority, BrownBoost and
RobustBoost.
\subsection{Boosting-by-majority}
Boost by majority (BBM) combines the base rules using equal weights for each
one of the rules. Two additional assumptions are made: that the error
of each of the base rules with respect to the corresponding
distribution is smaller than a fixed number: $1/2-\gamma$, and that
the number of boosting iterations is known in advance. Combining these
three restrictions allowed Freund to cast the learning problem in the
form of a mathematical game and find the optimal solution for that
game. The result is a potential function which depends both on the
margin and on the number of steps remaining until the end of the game.
Specifically, let $T$ be the total number of iterations and let $t$ 
be the current iteration. Suppose the example we currently consider is
$(x,y)$ and that $i$ is the number of existing base rules that predict
correctly on the $(x,y)$ ($i$ corresponds to the margin) then the
optimal boosting algorithm uses a potential function $\Phi_i^t$ that
is defined by the following recursion.
\begin{eqnarray} 
  \label{eqn:bbmPotentialInitial}
  \Phi_{i}^{T+1} = \cases{1 & if $i < 0$ \cr 0 & otherwise}\\
  \label{eqn:bbmPotential}
  \Phi_i^t = (\frac{1}{2} + \gamma) \Phi_{i+1}^{t+1} + (\frac{1}{2} - \gamma)\Phi_{i-1}^{t+1}
\end{eqnarray}

\subsection{BrownBoost and RobustBoost}
While BBM is a theoretically optimal boosting algorithm, it is not
applicable in practice, because it requires knowing the number of
steps in advance and giving each base rule the same weight. Contrast
this with Adaboost and LogitBoost which adapt their step size to the
error of the last base classifier.

Brownboost~\cite{Freund2001} overcomes this deficiency by taking the
limit of the BBM game where the number of steps goes to
infinity. Taking this limit is not trivial but the end result is
rather simple. In this limit both the margin $s$ and the time $t$ are
continuous:
\begin{equation} \label{eqn:brown-pot-closed}
\rpot{\s}{t}=\half \paren{
1-\erf \paren{ \frac{\s+2\sqrt{\bc}(1-t)}{\sqrt{2(1-t)}}}
}
\end{equation}

A slightly different limit yields the Robust-Boost potential:
\begin{equation} \label{eqn:RB-pot}
\rpot{\s}{t}=
\half \paren{
1-\erf \paren{ \frac{\s - \mu(t)}{\sigma(t)}}
}
\end{equation}
Where
\[ \sigma(t) = \sqrt{c_1 e^{-2t} - 1} \]
\[ \mu(t) = c_2 e^{-t} + 2\rho \] and $c_1,c_2$ are real valued
constants.

\subsection{Solving the potential function}
RobustBoost and BrownBoost require an additional step in the boosting
algorithm. The potential functions here change as a function of time,
and time is a continuous variable (it is not proportional to the
number of iterations).

We therefor need to solve at each iteration, a set of two non-linear
equations in two unknowns: the base rule weight $\alpha$ and the time
advance $\Delta t$. We use a standard numerical solver to do that.

\subsection{Setting the parameters}
Unlike Adaboost and Logitboost, The non-convex boosting algorithms
require choosing two parameters. These come in different forms, but
they are all equivalent to choosing the error goal $\epsilon$ and the
margin goal $\theta$. The error goal corresponds to a guess of the
fraction of the examples on which we need to ``give up''. While the
margin goal defines the minimal margin for the examples on which we
are not giving up. In the next section we propose an adaptive
algorithm for choosing $\epsilon$.

The time associated with the $n$-th iteration of of the BrownBoost and
RobustBoost algorithms is defined as $t_n=\sum_{i=1}^n (\Delta
t)_i$. The initial time equals zero $t_0=0$ and $t_{i-1} \geq t_i$
increases at each iteration. The termination time is defined to be
$t=1$. If the algorithm reaches that time it stops. In some cases the
setting for $\epsilon$ is too low and the setting for $\theta$ is too
high. As a result algorithm is is not able to reach $t=1$ even after a
large number of iterations. The final time reached by the algorithm is
a good indication of whether the parameters were set ambitiously,
causing the algorithm to never reach $t=1$ or not ambitiously enough,
causing $t=1$ to be reached after a small number of iterations.

We can use this as a method for tuning $\epsilon$ and $\theta$, but it
is a very slow process as each trial requires running the boosting
algorithm until it terminates or gets ``stuck''.

\section{Adaptive-$\epsilon$ Heuristic}
\begin{figure}[ht]
\centering
\includegraphics[trim=1.5cm 0cm 0cm 0cm, clip, width=0.9\textwidth]{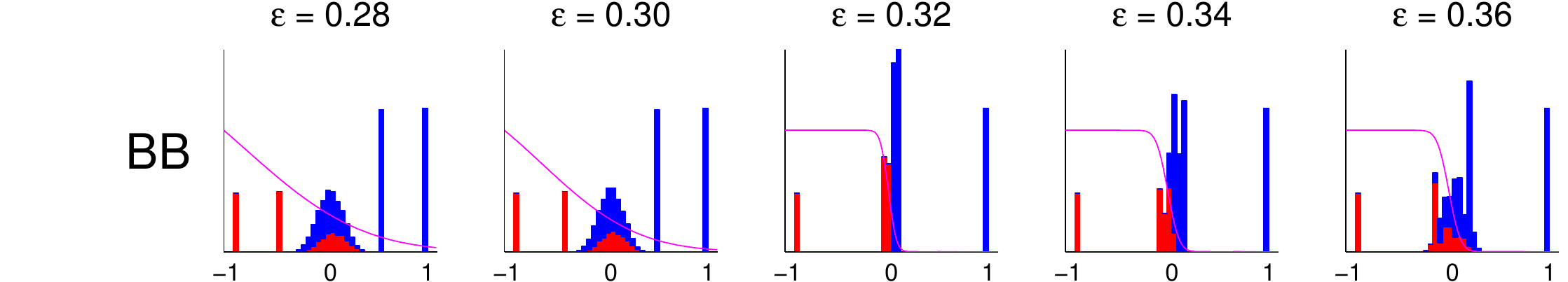}\\
\includegraphics[trim=1.5cm 0cm 0cm 0cm, clip, width=0.9\textwidth]{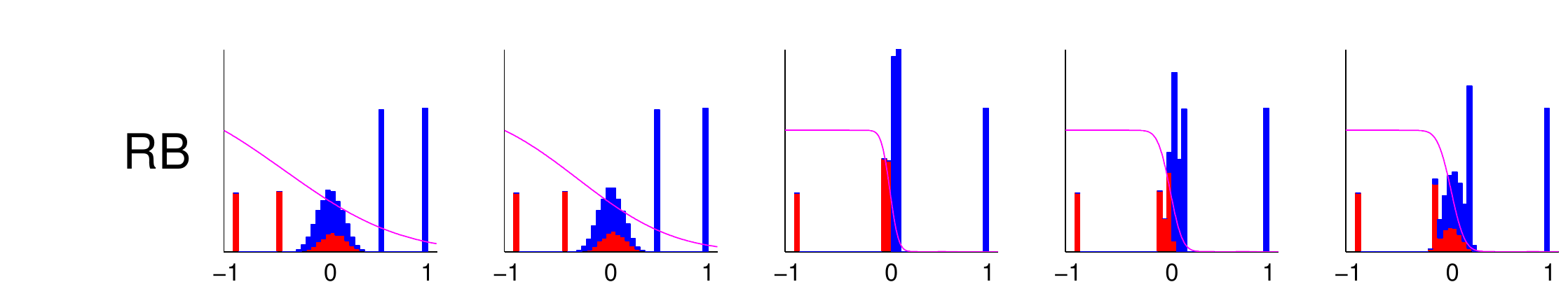}
\caption[Margin distributions of BrownBoost and RobustBoost using
different $\epsilon$ on Long dataset]{Margin distributions of
  BrownBoost (BB) and RobustBoost (RB) using different $\epsilon$ on
  LS dataset with 30\% label noise after 200 iterations. Noisy and
  clean examples are shown in red and blue respectively.}
\label{fig:long_bb_rb}
\end{figure}

In BrownBoost (BB) and RobustBoost (RB), we need to specify the target
error rate $\epsilon$. The choice of $\epsilon$ can greatly influence
the performance of the trained classifier. When $\epsilon$ is set too
low or too high, the algorithms often produce a classifier that
performs poorly even on the training data. Figure~\ref{fig:long_bb_rb}
shows the margin distributions of BB and RB using different $\epsilon$
on a dataset with 30\% label noise after 200 iterations. Note that
when $\epsilon \le 0.30$, the classifier cannot separate examples
around zero-margin.

When the true noise rate is known, a good rule of thumb is to set
$\epsilon$ a little higher than the noise
rate. Table~\ref{tab:boosting_info} summarizes the final time $t_f$
and the final training error rate $E_f$ of BB and RB on a synthetic
dataset with the true noise rate $\eta = \{0.1,0.2,0.3\}$ using two
different $\epsilon$ settings slightly above and below
$\eta$. However, when the true noise rate in not known, the tuning of
$\epsilon$ is usually done by cross validation. The process can be
very inefficient and usually involves a grid search over a small
interval.

\begin{table}[t]
\centering
\resizebox{0.75\textwidth}{!}{%
\begin{tabular}{ll|c|c|c|c|c|c|}
  \cline{3-8}
  &  & \multicolumn{2}{c|}{$\eta = 0.10$} & \multicolumn{2}{c|}{$\eta = 0.20$} & \multicolumn{2}{c|}{$\eta = 0.30$} \\ \cline{3-8} 
  &  & $\epsilon=\eta-0.02$ & $\epsilon=\eta+0.02$ & $\epsilon=\eta-0.02$ & $\epsilon=\eta+0.02$ & $\epsilon=\eta-0.02$ & $\epsilon=\eta+0.02$ \\ \hline
  \multicolumn{1}{|l|}{\multirow{2}{*}{$t_{f}$}} & BB & 0.19 & 0.30 & 0.22 & 0.81 & 0.25 & 0.82 \\ \cline{2-8} 
  \multicolumn{1}{|l|}{} & RB & 0.21 & 0.57 & 0.25 & 0.78 & 0.27 & 0.81 \\ \hline
  \multicolumn{1}{|l|}{\multirow{2}{*}{$E_{f}$}} & BB & 0.40 & 0.14 & 0.34 & 0.00 & 0.28 & 0.00 \\ \cline{2-8} 
  \multicolumn{1}{|l|}{} & RB & 0.40 & 0.00 & 0.34 & 0.00 & 0.28 & 0.00 \\ \hline
  \multicolumn{1}{|l|}{\multirow{2}{*}{$\frac{\vec{\alpha}\cdot\vec{h}}{||\vec{\alpha}||\cdot
        ||\vec{h}||}$}} & BB & 0.51 & 0.30 & 0.48 & 0.09 & 0.53 & 0.00 \\ \cline{2-8} 
  \multicolumn{1}{|l|}{} & RB & 0.51 & 0.11 & 0.48 & 0.08 & 0.53 & 0.00 \\ \hline
\end{tabular}
}
\caption{The final time $t_f$, the final training error rate
  with respect to clean labels $E_f$, and the angle between the true hypothesis $\vec{h}$ and
  $\vec{alpha}$ of BB and RB using $\epsilon$ slightly below and above
  the noise level $\eta$.
}
\label{tab:boosting_info}
\end{table}

In this paper we propose a heuristic for tuning $\epsilon$
automatically for BB and RB. The idea is based on the following
observation. When $\epsilon$ is too small, the time $t_k$ advances too
slowly that the boosting procedure seems ``stuck''. This situation can
often be remedied by slightly increasing the value of $\epsilon$
without having to restart the boosting process. The heuristic can be
described as follows. Initially, we set $\epsilon = 0$ and start the
boosting procedure. When the boosting algorithm does not to advance
for a few iterations or the numerical solver fails to solve the
non-linear equations, we slightly increase $\epsilon$ and resume the
boosting process. In our experiments, we denote BrownBoost and Robust
with adaptive-$\epsilon$ heuristic and RobustBoost with
adaptive-$\epsilon$, BBA and RBA respectively. 

\section{Experiments~\label{sec:experiments}}

In this section we compare the performance of BrownBoost with
adaptive-$\epsilon$ (BBA), RobustBoost with adaptive-$\epsilon$ (RBA)
to AdaBoost (ADB) and LogLossBoost (LLB)~\footnote{LogLossBoost is our
  implementation of LogitBoost with decision stumps.} on 3 datasets
with and without random label noise. Specifically we will look closely
at the margin distributions for each of the boosting
algorithms. Additionally, we will study the impact of using positive
target margin $\theta$ on BBA and RBA.

We implemented all of the boosting algorithms in MATLAB~\ and utilized
the Optimization Toolbox for numerically solving BB and RB
equations. For RBA, we use $\sigma_f = 0.001$ for all experiments.

\subsection{Datasets}
We conducted experiments on 3 different datasets: LS, Face and
Satimage. Each dataset can be described as follows. First, LS dataset
is a synthetic dataset whose construction is suggested by Long and
Servidio in~\cite{Long2008}. The dataset has input $x \in
\mathbb{R}^{21}$ with binary features $x_i \in {-1, +1}$ and label $y
\in {-1, +1}$. Each instance is generated as follows. First, the label
$y$ is chosen to be $-1$ or $+1$ with equal probability. Given $y$
and the margin width parameter $\delta$, the features $x_i$ are chosen
according to the following mixture distribution:
\begin{itemize}
\item \textbf{Large margin:} With probability $1/4$, we choose $x_i =
  y$ for all $1 \le i \le 21$
\item \textbf{Pullers:} With probability $1/4$, we choose $x_i = y$
  for $1 \le i \le 10+\delta$ and $x_i = -y$ for $11+\delta \le i \le
  21$
\item \textbf{Penalizers:} With probability $1/2$, we choose
  $5+\floor{\delta/2}$ random coordinates from the first 11 and
  $5+\ceil{\delta/2}$ from the last 10 to be equal to the label
  $y$. The remaining 10 coordinates are equal to $-y$.
\end{itemize} 
The data from this distribution can be classified perfectly by a
simple linear classifier $f(x) = sgn(\sum_i x_i)$. Note that $\delta$
essentially controls the separation margin of the examples. Larger
$\delta$ yields a larger margin.  

Face dataset is a collection of face and non-face images consisting of
10000 face images and 20000 non-faces images. For each image, a
feature vector of $\mathbb{R}^{176}$ is calculated based on histogram
of colors and gradients. When label noise is added to Face dataset, we only
added noise to the negative examples. We use 70\% of the examples for
training and 20\% for testing. 

Satimage is a dataset from the UCI repository~\cite{UCI}. There are
6435 examples and 36 attributes. The original label of 1-3 is grouped
as +1 and the rest is group as -1. Similar to Face dataset, we use
70\% of the examples for training and 20\% for testing.
 
\subsection{Results}
We first compared the performance of ADB, LLB, BBA and RBA with
different label noise level $\eta \in {0.0, 0.1, 0.2, 0.3}$ on LS
using 2 settings of the margin width parameter $\delta \in {1, 3}$. We
used the training set size $N = 1600$ and ran each boosting algorithm
for 200 iterations. For BBA and RBA, the margin parameter $\theta = 0$
and $\sigma_f = 0.001$.

When there was no label noise, all boosting algorithms managed to
learn the correct linear classifier. However, with presence of label
noise in both settings of $\delta$, BBA and RBA successfully converged
to the correct classifier while ADA and LLB did not as indicated by
the higher test error rates with respect to the true
labels. Table~\ref{tab:error_rates} summarizes the average test error
rates with respect to the true labels and the noisy labels over 10
runs and the standard deviation is reported in parentheses.

\begin{table}[t]
\centering
\resizebox{0.45\textwidth}{!}{
\begin{tabular}{ll|c|c|c|c|}
\cline{3-6} LS w/ $\delta=1$ & & ADB  & LLB  & BBA   & RBA   \\ 
\hline
\multicolumn{1}{|c}{\multirow{2}{*}{$\eta = 0.0$}} & \multicolumn{1}{|l|}{n} &
- & - & - & - \\ 
\cline{2-6} 
\multicolumn{1}{|l}{} & \multicolumn{1}{|l|}{t} & 0.00 (0.00) & 0.00 (0.00)& 0.00 (0.00)& 0.00 (0.00)\\ \hline
\multicolumn{1}{|c}{\multirow{2}{*}{$\eta = 0.1$}} & \multicolumn{1}{|l|}{n} &
0.25 (0.01) & 0.24 (0.01) & 0.10 (0.01)& 0.10 (0.01) \\ 
\cline{2-6} 
\multicolumn{1}{|l}{}  & \multicolumn{1}{|l|}{t} & 0.23 (0.01) & 0.22
(0.01) & 0.00 (0.00) & 0.01 (0.01) \\ \hline
\multicolumn{1}{|c}{\multirow{2}{*}{$\eta = 0.2$}} & \multicolumn{1}{|l|}{n} &
0.32 (0.01) & 0.31 (0.01) & 0.21 (0.01) & 0.22 (0.02) \\ 
\cline{2-6} 
\multicolumn{1}{|l}{} & \multicolumn{1}{|l|}{t} & 0.23 (0.01)
& 0.23 (0.01) & 0.03 (0.02) & 0.05 (0.03) \\ \hline
\multicolumn{1}{|c}{\multirow{2}{*}{$\eta = 0.3$}} & \multicolumn{1}{|l|}{n} &
0.36 (0.01) & 0.36 (0.01) & 0.31 (0.02) & 0.32 (0.02) \\ 
\cline{2-6} 
\multicolumn{1}{|l}{}  & \multicolumn{1}{|l|}{t} & 0.24 (0.01)
& 0.24 (0.01) & 0.09 (0.05) & 0.12 (0.05) \\ \hline
\end{tabular}
}
\hspace{1cm}
\resizebox{0.45\textwidth}{!}{
\begin{tabular}{ll|c|c|c|c|}
\cline{3-6} LS w/ $\delta=3$ &  & ADB  & LLB  & BBA   & RBA   \\ 
\hline
\multicolumn{1}{|c}{\multirow{2}{*}{$\eta = 0.0$}} & \multicolumn{1}{|l|}{n} &
- & - & - & - \\ 
\cline{2-6} 
\multicolumn{1}{|l}{} & \multicolumn{1}{|l|}{t} & 
0.00 (0.00) & 0.00 (0.00)& 0.00 (0.00)& 0.00 (0.00)\\ \hline
\multicolumn{1}{|c}{\multirow{2}{*}{$\eta = 0.1$}} & \multicolumn{1}{|l|}{n} &
0.11 (0.01) & 0.10 (0.01) & 0.10 (0.01) & 0.10 (0.01)\\ 
\cline{2-6} 
\multicolumn{1}{|l}{}  & \multicolumn{1}{|l|}{t} & 
0.02 (0.01) & 0.00 (0.00) & 0.01 (0.00) & 0.00 (0.00) \\ \hline
\multicolumn{1}{|c}{\multirow{2}{*}{$\eta = 0.2$}} & \multicolumn{1}{|l|}{n} &
0.22 (0.01) & 0.21 (0.01) & 0.19 (0.01) & 0.19 (0.01)  \\ 
\cline{2-6} 
\multicolumn{1}{|l}{} & \multicolumn{1}{|l|}{t} & 
0.06 (0.02) & 0.04 (0.02) & 0.02 (0.00) & 0.01 (0.01) \\ \hline
\multicolumn{1}{|c}{\multirow{2}{*}{$\eta = 0.3$}} & \multicolumn{1}{|l|}{n} &
0.33 (0.01) & 0.33 (0.01) & 0.29 (0.01) & 0.29 (0.01) \\ 
\cline{2-6} 
\multicolumn{1}{|l}{}  & \multicolumn{1}{|l|}{t} & 
0.12 (0.02) & 0.11 (0.02) & 0.04 (0.01) & 0.03 (0.01) \\ \hline
\end{tabular}
}
\caption{Average test error rates of ADB, LLB, BBA and RBA
  with respect to the noisy labels (n) and the true labels
  (t) on LS dataset with N=1600 in different noise settings.}
\label{tab:error_rates}
\end{table}

\begin{figure}[ht]
\centering
\subfloat[LS w/ $\delta = 1$]{
\includegraphics[trim=1.0cm 0cm 0cm 0cm, clip, width=0.5\textwidth]{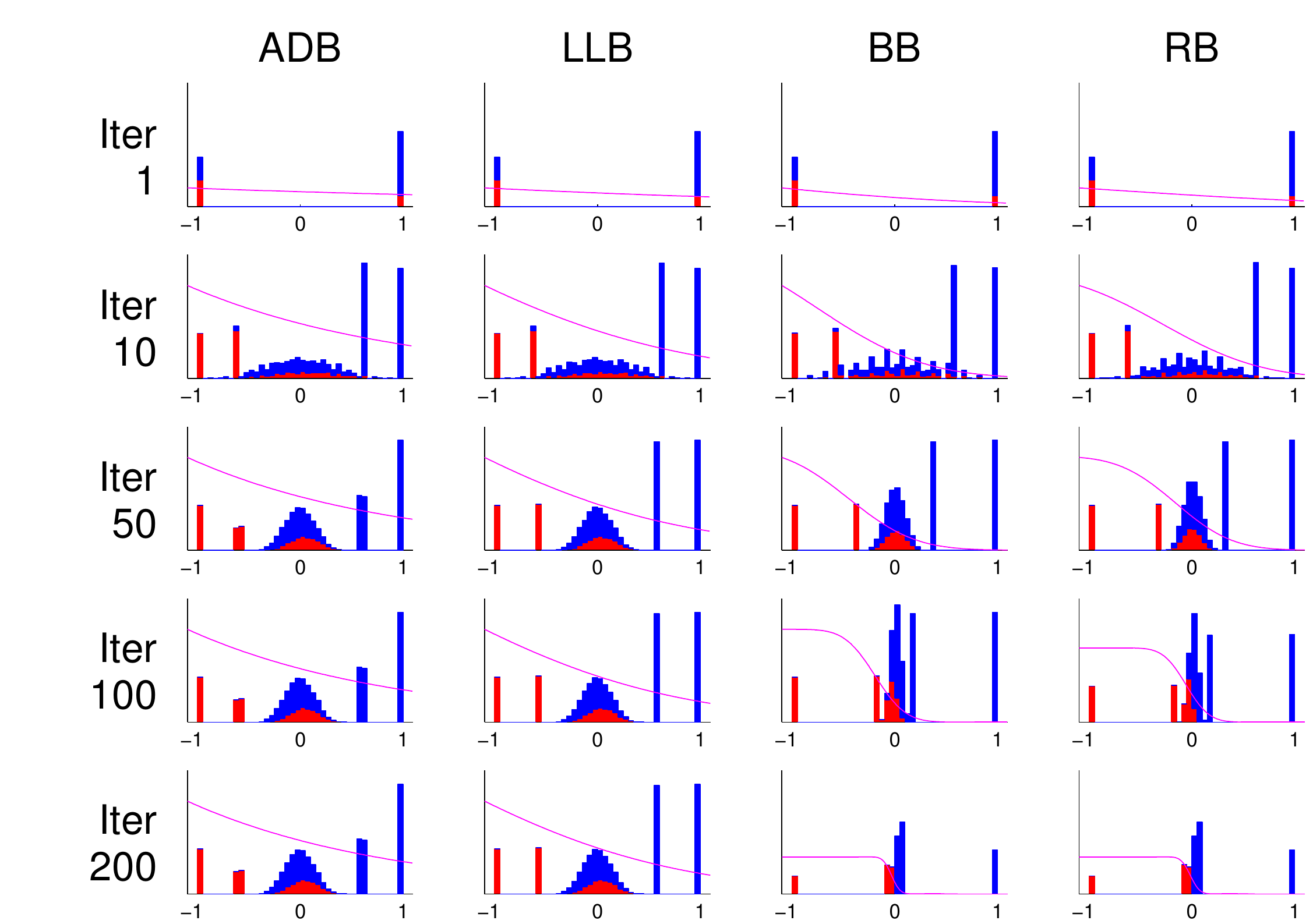}
}
\subfloat[LS w/ $\delta = 3$]{
\includegraphics[trim=1.0cm 0cm 0cm 0cm, clip, width=0.5\textwidth]{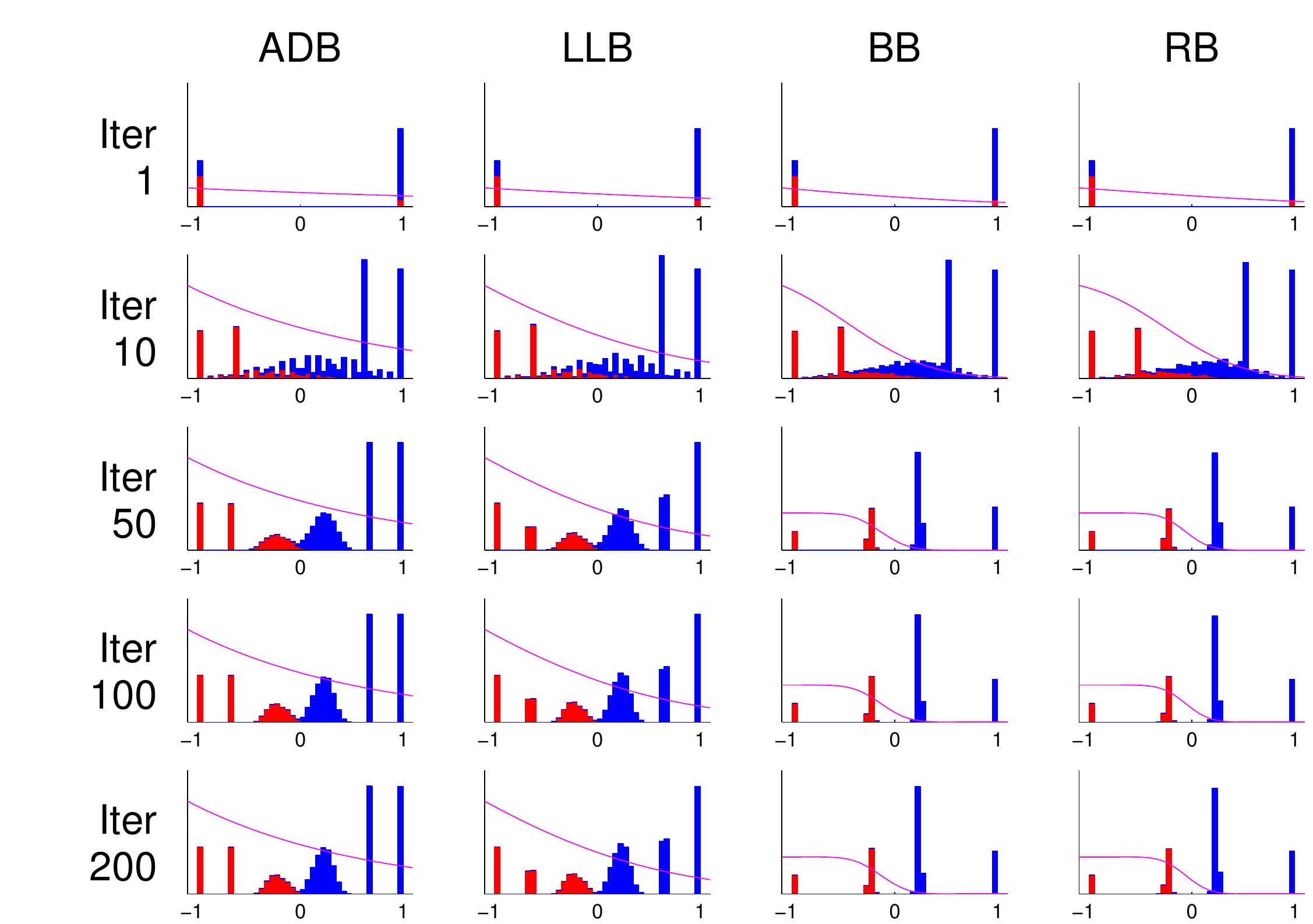}
}
\caption{Margin distribution progression and potential
  loss function of ADB, LLB, BBA
  and RBA on LS with 30\% label noise. Noisy and clean
  examples are shown in red and blue respectively. }
\label{fig:long_30_percent_noise}
\end{figure}

We also examined the progression of the margin distributions for each
of the boosting algorithms. Figure~\ref{fig:long_30_percent_noise}
shows the margin distributions progression when $\eta=0.3$. In LS with
$\delta=1$, ADB and LLB stopped progressing after 50 iterations due to
the large weights put on the noisy large-margin examples pushing the
classifier away from the correct hypothesis. On the contrary, after
100 iterations, RBA and BBA significantly decreased the weights of the
noisy large-margin examples as these examples are being ``given
up''. As a result, the boosting process continued on and eventually
converged to the correct classifier. Interestingly, in LS with $\delta
= 3$, we found that all boosting algorithms managed to attain the
training error rate of 0 in all noise settings. However, the test
error rates of ADB and LLB remained relatively high compared to those
of BBA and RBA.

\begin{figure}[t]
\centering
\includegraphics[width=0.8\textwidth]{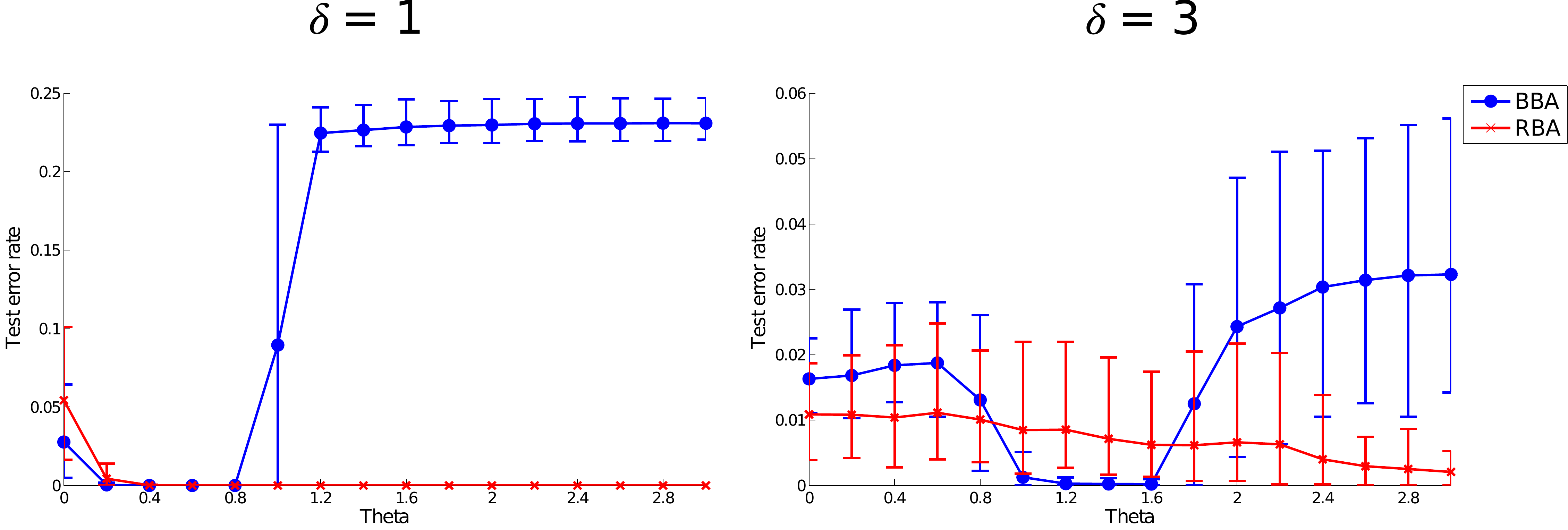}
\caption{Test errors of RBA and BBA using different $\theta$ on LS
  with 20\% noise. The whiskers indicate the minimum and maximum
  values over 10 runs.}
\label{fig:theta_long}
\end{figure}

\begin{figure}[t]
\centering
\includegraphics[width=0.8\textwidth]{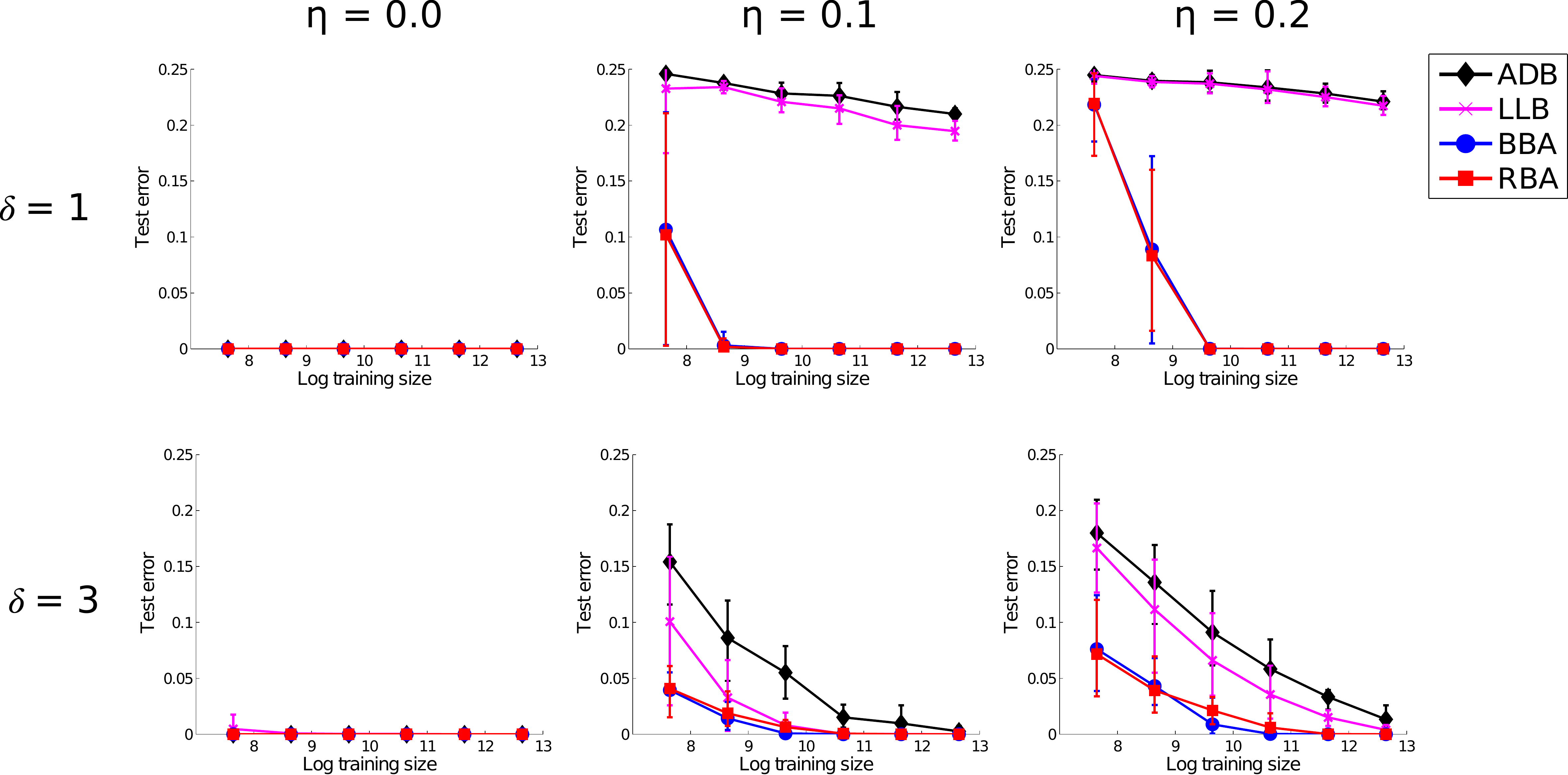}
\caption{Test errors of ADB, LLB, BBA, RBA while varying number of
  training examples at different noise levels $\eta$. The whiskers
  indicate the minimum and maximum values over 10 runs.}
\label{fig:long_summary_plot}
\end{figure}

We further explored the benefits of the margin parameter $\theta$. We
found that using a positive $\theta$ can improve generalization
error. Figure~\ref{fig:theta_long} shows the test errors of BBA and
RBA using different $\theta$ on LS with 20\% noise. Both algorithms
have lower generalization error when using positive $\theta$. We found
that RBA is less sensitive to the setting of $\theta$ than
BBA. Figure~\ref{fig:long_summary_plot} summarizes the test error
rates as a function of training set size for LS dataset with $\delta =
1$ and $\delta = 3$. For BBA and RBA, $\theta$ is tuned by 
cross-validation.

We also compared the performance of ADB, LLB, BBA and RBA on Face and
Satimage. We ran each boosting algorithm for 800 iterations on both
datasets with 2 different noise levels $\eta = {0.0, 0.2}$. We also
repeated the experiment after holding out of 75\% of the training
examples. The area under the average ROC curves is summarized in
Table~\ref{tab:AUC}.

\begin{table}[t]
\centering
\subfloat[Face]{
\resizebox{0.9\textwidth}{!}{
\begin{tabular}{|l|c|c|c|c|}
\hline
      & ADB             & LLB             & BBA             & RBA             \\ \hline
$\eta=0.0, N=21000$  & 0.9996 (0.0001) & 0.9979 (0.0004) & 0.9996
(0.0001) & 0.9996 (0.0000) \\
\hline
$\eta=0.0, N=5250$  & 0.9998 (0.0000) & 0.9994 (0.0001) & 0.9998
(0.0000) & 0.9997 (0.0000) \\
\hline
$\eta=0.2, N=21000$  & 0.9991 (0.0001) & 0.9992 (0.0001) & 0.9995
(0.0001) & 0.9995 (0.0001) \\
\hline
$\eta=0.2, N=5250$ & 0.9983 (0.0003) & 0.9982 (0.0003) & 0.9992
(0.0001) & 0.9993 (0.0001) \\
\hline
\end{tabular}
}
}\\
\subfloat[Satimage]{
\resizebox{0.9\textwidth}{!}{
\begin{tabular}{|l|c|c|c|c|}
\hline
      & ADB             & LLB             & BBA             & RBA             \\ \hline
$\eta=0.0, N=4504$  & 0.9830 (0.0003) & 0.9832 (0.0007) & 0.9770
(0.0010) & 0.9757 (0.0005) \\
\hline
$\eta=0.0, N=1126$  & 0.9764 (0.0018) & 0.9720 (0.0021) & 0.9770
(0.0016) & 0.9729 (0.0022) \\
\hline
$\eta=0.2, N=4504$  & 0.9679 (0.0026) & 0.9699 (0.0021) & 0.9749
(0.0018) & 0.9715 (0.0038) \\
\hline
$\eta=0.2, N=1126$ & 0.9459 (0.0043) & 0.9421 (0.0053) & 0.9607
(0.0039) & 0.9656 (0.0047) \\
\hline
\end{tabular}
}
}
\caption{Average area under ROC}
\label{tab:AUC}
\end{table}

For both Face and Satimage, using paired t-test, we found that area under ROC of
RBA and BBA is significantly larger than that of LLB with $p < 0.001$
in all settings. We also found that the difference between ADB and RBA
is insignificant in noise-free cases and the difference between RBA
and BBA is insignificant in all cases. For Satimage dataset,

\section{Conclusion} \label{sec:conclusion}
Our experiments show that Brownboost and Robustboost are significantly
more resistant to label noise than Adaboost and LogitBoost. We show
how this is related to the progressions of the margin distribution
over time. We show that the setting of the target error rate
$\epsilon$ is of critical importance for the final performance and
provide a practical heuristics for setting it. Our experiments also
show that, for noisy small training sets, maximizing the margin on the
examples on which we don't ``give up'' is significantly better than
minimizing the training error.

\clearpage
\bibliographystyle{abbrv}
\bibliography{sunsern_nips2014}

\end{document}